\documentclass[11pt,a4paper]{article}
\usepackage{authblk}
\usepackage[hyperref]{acl2017}
\usepackage{times}
\usepackage{latexsym}
\usepackage{amsmath}
\usepackage{url}
\usepackage[pdftex]{graphicx}
\usepackage{caption}

\aclfinalcopy % Uncomment this line for the final submission
\def\aclpaperid{429} %  Enter the acl Paper ID here

\title{Affect-LM: A Neural Language Model for Customizable Affective Text Generation}

\author[1]{Sayan Ghosh}
\author[1]{Mathieu Chollet}
\author[1]{Eugene Laksana}
\author[2]{Louis-Philippe Morency}
\author[1]{Stefan Scherer}
\affil[1]{Institute for Creative Technologies, University of Southern California, CA, USA}
\affil[2]{Language Technologies Institute, Carnegie Mellon University, PA, USA}
\affil[1]{\textit {\{sghosh,chollet,elaksana,scherer\}@ict.usc.edu}} \affil[2]{\textit{morency@cs.cmu.edu}}

\date{}

\begin{document}
\maketitle
\begin{abstract}
 Human verbal communication includes affective messages which are conveyed through use of emotionally colored words. There has been a lot of research in this direction but the problem of integrating state-of-the-art neural language models with affective information remains an area ripe for exploration. In this paper, we propose an extension to an LSTM (Long Short-Term Memory) language model for generating conversational text, conditioned on affect categories. Our proposed model, \textit{Affect-LM} enables us to customize the degree of emotional content in generated sentences through an additional design parameter. Perception studies conducted using Amazon Mechanical Turk show that \textit{Affect-LM} generates naturally looking emotional sentences without sacrificing grammatical correctness. \textit{Affect-LM} also learns affect-discriminative word representations, and perplexity experiments show that additional affective information in conversational text can improve language model prediction.  
\end{abstract}

\section{Introduction}
\label{sec:intro}
Affect is a term that subsumes emotion and longer term constructs such as mood and personality and refers to the experience of feeling or emotion~\cite{scherer2010blueprint}.~\citeauthor{picard1997affective}~\shortcite{picard1997affective} provides a detailed discussion of the importance of affect analysis in human communication and interaction. Within this context the analysis of human affect from text is an important topic in natural language understanding, examples of which include sentiment analysis from Twitter~\cite{nakov2016semeval}, affect analysis from poetry~\cite{kao2012computational} and studies of correlation between function words and social/psychological processes~\cite{pennebaker2011secret}. People exchange verbal messages which not only contain syntactic information, but also information conveying their mental and emotional states. Examples include the use of emotionally colored words (such as \textit{furious} and \textit{joy}) and swear words. The automated processing of affect in human verbal communication is of great importance to understanding spoken language systems, particularly for emerging applications such as dialogue systems and conversational agents. 

\begin{figure}[t]
 \centering
 \captionsetup{justification=centering}
  \includegraphics[scale=0.50]{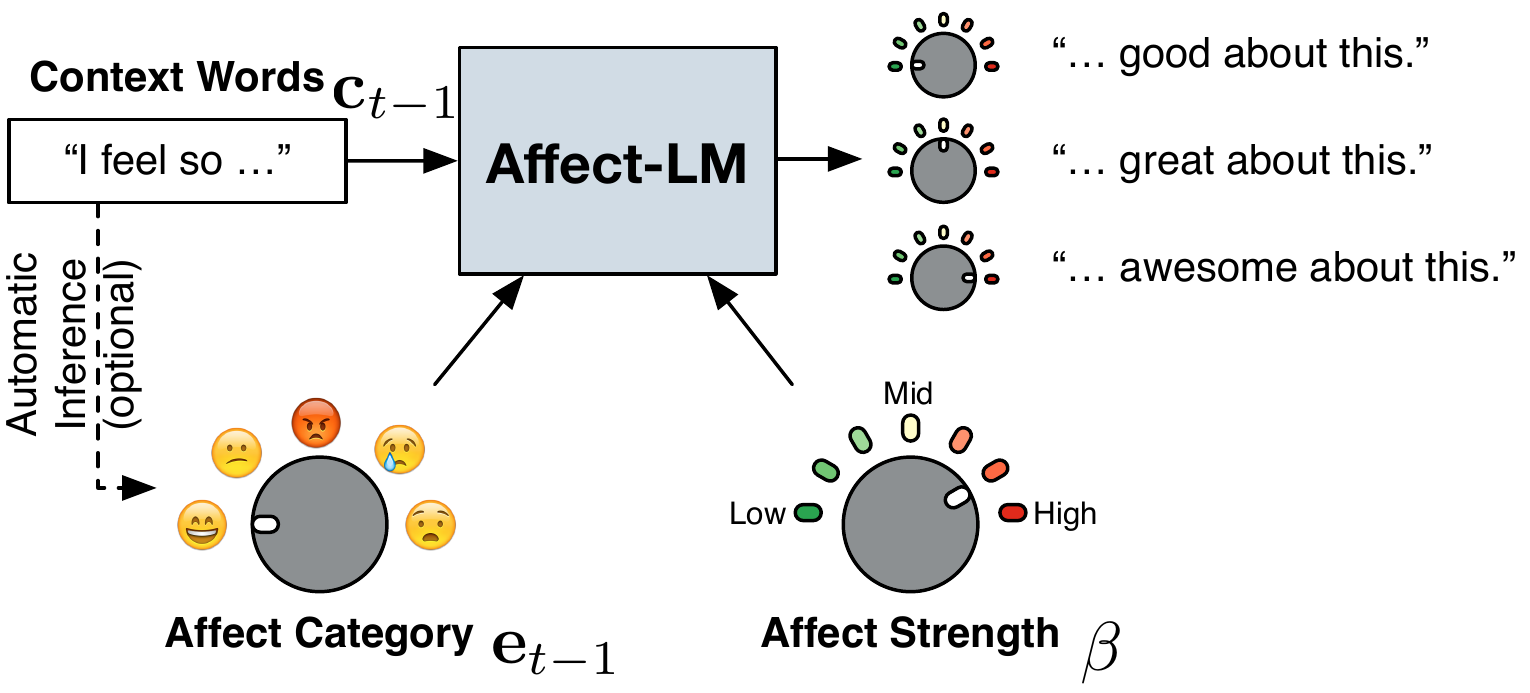}
  \caption{\emph{Affect-LM} is capable of generating emotionally colored conversational text in five specific affect categories ($\mathbf{e}_{t-1}$) with varying affect strengths ($\beta$). Three generated example sentences for \emph{happy} affect category are shown in three distinct affect strengths.}
	\label{overview-fig}
\end{figure}

Statistical language modeling is an integral component of speech recognition systems, with other applications such as machine translation and information retrieval. There has been a resurgence of research effort in recurrent neural networks for language modeling~\cite{mikolov2010recurrent}, which have yielded performances far superior to baseline language models based on n-gram approaches. However, there has not been much effort in building neural language models of text that leverage affective information. Current literature on deep learning for language understanding focuses mainly on representations based on word semantics~\cite{mikolov2013distributed}, encoder-decoder models for sentence representations~\cite{cho2015describing}, language modeling integrated with symbolic knowledge~\cite{ahn2016neural} and neural caption generation~\cite{Vinyals_2015_CVPR}, but to the best of our knowledge there has been no work on augmenting neural language modeling with affective information, or on data-driven approaches to generate emotional text. 

Motivated by these advances in neural language modeling and affective analysis of text, in this paper we propose a model for representation and generation of emotional text, which we call the \textit{Affect-LM}. Our model is trained on conversational speech corpora, common in language modeling for speech recognition applications~\cite{bulyko2007web}. Figure~\ref{overview-fig} provides an overview of our \emph{Affect-LM} and its ability to generate emotionally colored conversational text in a number of affect categories with varying affect strengths. While these parameters can be manually tuned to generate conversational text, the affect category can also be automatically inferred from preceding context words. Specifically for model training, the affect category is derived from features generated using keyword spotting from a dictionary of emotional words, such as the LIWC (Linguistic Inquiry and Word Count) tool~\cite{pennebaker2001linguistic}. Our primary research questions in this paper are: \\ 
\textbf{Q1:}Can \textit{Affect-LM} be used to generate affective sentences for a target emotion with varying degrees of affect strength through a customizable model parameter? \\ 
\textbf{Q2:}Are these generated sentences rated as emotionally expressive as well as grammatically correct in an extensive crowd-sourced perception experiment? \\
\textbf{Q3:}Does the automatic inference of affect category from the context words improve language modeling performance of the proposed \textit{Affect-LM} over the baseline as measured by perplexity? 

The remainder of this paper is organized as follows. In Section~\ref{sec:prior-work}, we discuss prior work in the fields of neural language modeling, and generation of affective conversational text. In Section~\ref{sec:model-description} we describe the baseline LSTM model and our proposed \textit{Affect-LM} model. Section~\ref{sec:expr-settings} details the experimental setup, and in Section~\ref{sec:results}, we discuss results for customizable emotional text generation, perception studies for each affect category, and perplexity improvements over the baseline model before concluding the paper in Section~\ref{sec:conclusions}.

\section{Related Work}
\label{sec:prior-work}
Language modeling is an integral component of spoken language systems, and traditionally n-gram approaches have been used ~\cite{stolcke2002srilm} with the shortcoming that they are unable to generalize to word sequences which are not in the training set, but are encountered in unseen data. ~\citeauthor{bengio2003neural} \shortcite{bengio2003neural} proposed neural language models, which address this shortcoming by generalizing through word representations. \citeauthor{mikolov2010recurrent}~\shortcite{mikolov2010recurrent} and \citeauthor{sundermeyer2012lstm}~\shortcite{sundermeyer2012lstm} extend neural language models to a recurrent architecture, where a target word $w_t$ is predicted from a context of all preceding words $w_1, w_2,..., w_{t-1}$ with an LSTM (Long Short-Term Memory) neural network. There also has been recent effort on building language models conditioned on other modalities or attributes of the data. For example, ~\citeauthor{Vinyals_2015_CVPR}~\shortcite{Vinyals_2015_CVPR} introduced the neural image caption generator, where representations learnt from an input image by a CNN (Convolutional Neural Network) are fed to an LSTM language model to generate image captions. ~\citeauthor{kiros2014multimodal}~\shortcite{kiros2014multimodal} used an LBL model (Log-Bilinear language model) for two applications - image retrieval given sentence queries, and image captioning. Lower perplexity was achieved on text conditioned on images rather than language models trained only on text. 

In contrast, previous literature on affective language generation has not focused sufficiently on customizable state-of-the-art neural network techniques to generate emotional text, nor have they quantitatively evaluated their models on multiple emotionally colored corpora.~\citeauthor{mahamood2011generating}~\shortcite{mahamood2011generating} use several NLG (natural language generation) strategies for producing affective medical reports for parents of neonatal infants undergoing healthcare. While they study the difference between affective and non-affective reports, their work is limited only to heuristic based systems and do not include conversational text.~\citeauthor{mairesse2007personage}~\shortcite{mairesse2007personage} developed PERSONAGE, a system for dialogue generation conditioned on extraversion dimensions. They trained regression models on ground truth judge's selections to automatically determine which of the sentences selected by their model exhibit appropriate extroversion attributes.
%~\citeauthor{malandrakis2014affective}~\shortcite{malandrakis2014affective} proposed a method to adapt affect lexicon generation to individual domains, however their method is tailored to lexicon generation rather than affective language generation.
In~\citeauthor{keshtkar2011pattern}~\shortcite{keshtkar2011pattern}, the authors use heuristics and rule-based approaches for emotional sentence generation. Their generation system is not training on large corpora and they use additional syntactic knowledge of parts of speech to create simple affective sentences. In contrast, our proposed approach builds on state-of-the-art approaches for neural language modeling, utilizes no syntactic prior knowledge, and generates expressive emotional text. 

\begin{table*}
\centering
 \captionsetup{justification=centering}
\scriptsize
\begin{tabular}{|l|c|c|c|c|}
\hline
{\bf Corpus Name} & {\bf Conversations} & {\bf Words} & {\bf \% Colored Words} & {\bf Content}\\\hline
Fisher & 11700 & 21167581 & 3.79 & Conversations 	\\	\hline 
DAIC & 688 & 677389 & 5.13 & Conversations \\	\hline
SEMAINE & 959  & 23706 & 6.55 & Conversations \\ \hline
CMU-MOSI & 93 & 26121 & 6.54 & Monologues \\ \hline
\end{tabular}
\caption{Summary of corpora used in this paper. CMU-MOSI and SEMAINE are observed to have higher emotional content than Fisher and DAIC corpora.}
\label{dataset summary}
\end{table*}

\section{Model}
\label{sec:model-description}
\subsection{LSTM Language Model}
Prior to providing a formulation for our proposed model, we briefly describe a LSTM language model. We have chosen this model as a baseline since it has been reported to achieve state-of-the-art perplexities compared to other approaches, such as n-gram models with Kneser-Ney smoothing~\cite{jozefowicz2016exploring}. Unlike an ordinary recurrent neural network, an LSTM network does not suffer from the vanishing gradient problem which is more pronounced for very long sequences~\cite{hochreiter1997long}. Formally, by the chain rule of probability, for a sequence of $M$ words $w_1, w_2,..., w_M$, the joint probability of all words is given by:
\begin{equation}
P(w_1, w_2,..., w_M) = \prod_{t=1}^{t=M} P(w_t|w_1, w_2,...., w_{t-1})
\end{equation}
If the vocabulary consists of $V$ words, the conditional probability of word $w_t$ as a function of its context $\mathbf{c_{t-1}}=(w_1, w_2,...., w_{t-1})$ is given by:
\begin{equation}
\label{baseline-eqn}
P(w_t=i|\mathbf{c_{t-1}})=\frac{\exp(\mathbf{U_i}^T\mathbf{f(c_{t-1})}+b_i)}{\sum_{i=1}^{V} \exp(\mathbf{U_i}^T\mathbf{f(c_{t-1})}+b_i)}
\end{equation}
$\mathbf{f(.)}$ is the output of an LSTM network which takes in the context words $w_1, w_2,...,w_{t-1}$ as inputs through one-hot representations, $\mathbf{U}$ is a matrix of word representations which on visualization we have found to correspond to POS (Part of Speech) information, while $\mathbf{b_i}$ is a bias term capturing the unigram occurrence of word $i$. Equation~\ref{baseline-eqn} expresses the word $w_t$ as a function of its context for a LSTM language model which does not utilize any additional affective information. 
\subsection{Proposed Model: \textit{Affect-LM}}
\label{sec:proposed-model}
The proposed model \textit{Affect-LM} has an additional energy term in the word prediction, and can be described by the following equation:
\begin{equation}
\label{affectlm-eqn}
\begin{split}
\small{P(w_t=i|\mathbf{c_{t-1}},\mathbf{e_{t-1}})= \qquad \qquad \qquad \qquad \qquad \qquad} \\
\small{\frac{\exp{ (\mathbf{U_i}^T\mathbf{f(c_{t-1})}+\beta\mathbf{V_i}^T\mathbf{g(e_{t-1})}+b_i) }}{\sum_{i=1}^{V} \exp(\mathbf{U_i}^T\mathbf{f(c_{t-1})}+\beta\mathbf{V_i}^T\mathbf{g(e_{t-1})}+b_i)}}
\end{split}
\end{equation}
$\mathbf{e_{t-1}}$ is an input vector which consists of affect category information obtained from the words in the context during training, and $\mathbf{g(.)}$ is the output of a network operating on $\mathbf{e_{t-1}}$.$\mathbf{V_i}$ is an embedding learnt by the model for the $i$-th word in the vocabulary and is expected to be discriminative of the affective information conveyed by each word. %For example, for the words \textit{anger} and \textit{sad}, we would expect their corresponding embeddings to reside in a region of the word space opposite to the embedding for the word \textit{joy}. 
In Figure~\ref{embed-repr} we present a visualization of these affective representations. \\
The parameter $\beta$ defined in Equation~\ref{affectlm-eqn}, which we call the \textit{affect strength} defines the influence of the affect category information (frequency of emotionally colored words) on the overall prediction of the target word $w_t$ given its context. We can consider the formulation as an energy based model (EBM), where the additional energy term captures the degree of correlation between the predicted word and the affective input~\cite{bengio2003neural}.
\subsection{Descriptors for Affect Category Information}
Our proposed model learns a generative model of the next word $w_t$ conditioned not only on the previous words $w_1,w_2,...,w_{t-1}$ but also on the \textit{affect category} $\mathbf{e_{t-1}}$ which is additional information about emotional content. During model training, the affect category is inferred from the context data itself. Thus we define a suitable feature extractor which can utilize an affective lexicon to \textit{infer} emotion in the context. For our experiments, we have utilized the Linguistic Inquiry and Word Count (LIWC) text analysis program for feature extraction through keyword spotting. Introduced by~\citeauthor{pennebaker2001linguistic} \shortcite{pennebaker2001linguistic}, LIWC is based on a dictionary, where each word is assigned to a predefined LIWC category. The categories are chosen based on their association with social, affective, and cognitive processes. For example, the dictionary word \textit{worry} is assigned to LIWC category \textit{anxiety}. %Prior literature has used LIWC for the affect and social analysis of text, with widespread applications such as depression detection~\cite{nguyen2014affective} and social media analysis~\cite{gilbert2009predicting}.
In our work, we have utilized all word categories of LIWC corresponding to affective processes: \textit{positive emotion, angry, sad, anxious}, and \textit{negative emotion}. Thus the descriptor $\mathbf{e_{t-1}}$ has five features with each feature denoting presence or absence of a specific emotion, which is obtained by binary thresholding of the features extracted from LIWC. For example, the affective representation of the sentence \textit{i will fight in the war} is $\mathbf{e_{t-1}}=$\{``sad":0, ``angry":1, ``anxiety":0, ``negative emotion":1, ``positive emotion":0\}. 
\subsection{\textit{Affect-LM} for Emotional Text Generation}
\label{subsec:generate-text}
\textit{Affect-LM} can be used to generate sentences conditioned on the input affect category, the affect strength $\beta$, and the context words. For our experiments, we have chosen the following affect categories - \emph{positive emotion, anger, sad, anxiety}, and \emph{negative emotion} (which is a superclass of anger, sad and anxiety). As described in Section~\ref{sec:proposed-model}, the affect strength $\beta$ defines the degree of dominance of the affect-dependent energy term on the word prediction in the language model, consequently after model training we can change $\beta$ to control the degree of how ``emotionally colored" a generated utterance is, varying from $\beta=0$ (neutral; baseline model) to $\beta=\infty$ (the generated sentences only consist of emotionally colored words, with no grammatical structure). %Figure~\ref{overview-fig} shows a summary of sentence generation for \textit{Affect-LM}, with the role of $\beta$ in controlling emotional strength. 

When \textit{Affect-LM} is used for generation, the affect categories could be either (1) inferred from the context using LIWC (this occurs when we provide sentence beginnings which are emotionally colored themselves), or (2) set to an input emotion descriptor $\mathbf{e}$ (this is obtained by setting $\mathbf{e}$ to a binary vector encoding the desired emotion and works even for neutral sentence beginnings). Given an initial starting set of $M$ words $w_1,w_2,...,w_M$ to complete, affect strength $\beta$, and the number of words $N$ to generate each $i$-th generated word is obtained by sampling from $P(w_i | w_1, w_2,...,w_{i-1},\mathbf{e};\beta)$ for $i \in \{M+1, M+2, ..., M+N\}$. 
\section{Experimental Setup}
\label{sec:expr-settings}
In Section~\ref{sec:intro}, we have introduced three primary research questions related to the ability of the proposed \textit{Affect-LM} model to generate emotionally colored conversational text without sacrificing grammatical correctness, and to obtain lower perplexity than a baseline LSTM language model when evaluated on emotionally colored corpora. In this section, we discuss our experimental setup to address these questions, with a description of \textit{Affect-LM}'s architecture and the corpora used for training and evaluating the language models.
\subsection{Speech Corpora}
\label{sec: speech-corpuses}
The Fisher English Training Speech Corpus is the main corpus used for training the proposed model, in addition to which we have chosen three emotionally colored conversational corpora. A brief description of each corpus is given below, and in Table~\ref{dataset summary}, we report relevant statistics, such as the total number of words, along with the fraction of emotionally colored words (those belonging to the LIWC affective word categories) in each corpus. \\
\textbf{Fisher English Training Speech Parts 1 \& 2:} The Fisher dataset~\cite{cieri2004fisher} consists of speech from telephonic conversations of 10 minutes each, along with their associated transcripts. Each conversation is between two strangers who are requested to speak on a randomly selected topic from a set. Examples of conversation topics are \textit{Minimum Wage}, \textit{Time Travel} and \textit{Comedy}.\\
\textbf{Distress Assessment Interview Corpus (DAIC)}: The DAIC corpus introduced by ~\citeauthor{gratch2014distress}~\shortcite{gratch2014distress} consists of 70+ hours of dyadic interviews between a human subject and a virtual human, where the virtual human asks questions designed to diagnose symptoms of psychological  distress in the subject such  as  depression  or  PTSD  (Post Traumatic Stress Disorder). \\
\textbf{SEMAINE dataset:} SEMAINE~\cite{mckeown2012semaine} is a large audiovisual corpus consisting of interactions between subjects and an operator simulating a SAL (Sensitive Artificial Listener). There are a total of 959 conversations which are approximately 5 minutes each, and are transcribed and annotated with affective dimensions. \\
\textbf{Multimodal Opinion-level Sentiment Intensity Dataset (CMU-MOSI):}~\cite{zadeh2016multimodal} This is a multimodal annotated corpus of opinion videos where in each video a speaker expresses his opinion on a commercial product. The corpus consist of speech from 93 videos from 89 distinct speakers (41 male and 48 female speakers). This corpus differs from the others since it contains monologues rather than conversations.

While we find that all corpora contain spoken language, they have the following characteristics different from the Fisher corpus: (1) More emotional content as observed in Table~\ref{dataset summary}, since they have been generated through a human subject's spontaneous replies to questions designed to generate an emotional response, or from conversations on emotion-inducing topics (2) Domain mismatch due to recording environment (for example, the DAIC corpus was created in a mental health setting, while the CMU-MOSI corpus consisted of opinion videos uploaded online). (3) Significantly smaller than the Fisher corpus, which is 25 times the size of the other corpora combined. Thus, we perform training in two separate stages - training of the baseline and \textit{Affect-LM} models on the Fisher corpus, and subsequent adaptation and fine-tuning on each of the emotionally colored corpora.

\subsection{\textit{Affect-LM} Neural Architecture}
For our experiments, we have implemented a baseline LSTM language model in Tensorflow~\cite{abadi2016tensorflow}, which follows the non-regularized implementation as described in~\citeauthor{zaremba2014recurrent}~\shortcite{zaremba2014recurrent} and to which we have added a separate energy term for the affect category in implementing \textit{Affect-LM}. We have used a vocabulary of 10000 words and an LSTM network with 2 hidden layers and 200 neurons per hidden layer. The network is unrolled for 20 time steps, and the size of each minibatch is 20. The affect category $\mathbf{e_{t-1}}$ is processed by a multi-layer perceptron with a single hidden layer of 100 neurons and sigmoid activation function to yield $\mathbf{g(e_{t-1})}$. We have set the output layer size to 200 for both $\mathbf{f(c_{t-1})}$ and $\mathbf{g(e_{t-1})}$. We have kept the network architecture constant throughout for ease of comparison between the baseline and \textit{Affect-LM}.

\begin{table*}
\centering
 \captionsetup{justification=centering}
\scriptsize
\begin{tabular}{|l|l|l|}
\hline
{\bf Beginning} & {\bf Affect Category} & {Completed sentence}\\\hline
\textit{\textbf{I feel so}} & Happy & good because i think that it's important to have a relationship with a friend \\
          & Angry & bad that i hate it and i hate that because they they kill themselves and then they fight \\
          & Sad & sad to miss because i i miss the feelings of family members who i lost feelings with \\ 
          & Anxious & horrible i mean i think when we're going to you know war and alert alert and we're actually gonna die \\ 
					& Neutral & bad if i didn't know that the decision was going on \\
					\hline 
\textit{\textbf{I told him to}} & Happy & be honest and i said well i hope that i 'm going to be a better person\\
          & Angry & see why he was fighting with my son\\
					& Sad & leave the house because i hurt one and i lost his leg and hurt him \\
					& Anxious & be afraid of him and he he just he just didn't care about the death penalty\\
					& Neutral & do this position i think he is he's got a lot of money he has to pay himself a lot of money\\
					\hline
\textit{\textbf{Why did you}} & Happy & have a best friend \\
                              & Angry & say it was only a criminal being killed at a war or something \\
															& Sad & miss your feelings \\
															& Anxious & worry about fear factor \\ 
															& Neutral & believe in divorce \\
					\hline
\end{tabular}
\caption{Example sentences generated by the model conditioned on different affect categories}\label{tab:sampling}
\end{table*}

\subsection{Language Modeling Experiments}
\label{sec:expr}
\textit{Affect-LM} can also be used as a language model where the next predicted word is estimated from the words in the context, along with an affect category extracted from the context words themselves (instead of being encoded externally as in generation). To evaluate whether additional emotional information could improve the prediction performance, we train the corpora detailed in Section~\ref{sec: speech-corpuses} in two stages as described below: \\
(1) \textbf{Training and validation of the language models on Fisher dataset}- The Fisher corpus is split in a 75:15:10 ratio corresponding to the training, validation and evaluation subsets respectively, and following the implementation in~\citeauthor{zaremba2014recurrent}~\shortcite{zaremba2014recurrent}, we train the language models (both the baseline and \textit{Affect-LM}) on the training split for 13 epochs, with a learning rate of 1.0 for the first four epochs, and the rate decreasing by a factor of 2 after every subsequent epoch. The learning rate and neural architecture are the same for all models. We validate the model over the affect strength $\beta \in [1.0, 1.5, 1.75, 2.0, 2.25, 2.5, 3.0]$. The best performing model on the Fisher validation set is chosen and used as a seed for subsequent adaptation on the emotionally colored corpora. \\
(2) \textbf{Fine-tuning the seed model on other corpora}- Each of the three corpora - CMU-MOSI, DAIC and SEMAINE are  split in a 75:15:10 ratio to create individual training, validation and evaluation subsets. For both the baseline and \textit{Affect-LM}, the best performing model from Stage 1 (the seed model) is fine-tuned on each of the training corpora, with a learning rate of 0.25 which is constant throughout, and a validation grid of $\beta \in [1.0, 1.5, 1.75, 2.0]$. For each model adapted on a corpus, we compare the perplexities obtained by \textit{Affect-LM} and the baseline model when evaluated on that corpus. 

\subsection{Sentence Generation Perception Study}
We assess \emph{Affect-LM's} ability to generate emotionally colored text of varying degrees without severely deteriorating grammatical correctness, by conducting an extensive perception study on Amazon's Mechanical Turk (MTurk) platform. The MTurk platform has been successfully used in the past for a wide range of perception experiments and has been shown to be an excellent resource to collect human ratings for large studies \cite{buhrmester2011amazon}. Specifically, we generated more than 200 sentences for four sentence beginnings (namely the three sentence beginnings listed in Table \ref{tab:sampling} as well as an end of sentence token indicating that the model should generate a new sentence) in five affect categories \textit{happy(positive emotion), angry, sad, anxiety}, and \textit{negative emotion}. The \textit{Affect-LM} model trained on the Fisher corpus was used for sentence generation. Each sentence was evaluated by two human raters that have a minimum approval rating of 98\% and are located in the United States. The human raters were instructed that the sentences should be considered to be taken from a conversational rather than a written context: repetitions and pause fillers (e.g., \emph{um, uh}) are common and no punctuation is provided. The human raters evaluated each sentence on a seven-point Likert scale for the five affect categories, overall \emph{affective valence} as well as the sentence's \emph{grammatical correctness} and were paid 0.05USD per sentence. We measured inter-rater agreement using Krippendorff’s $\alpha$ and observed considerable agreement between raters across all categories (e.g., for \emph{valence} $\alpha = 0.510$ and \emph{grammatical correctness} $\alpha = 0.505$).

For each \emph{target emotion} (i.e., intended emotion of generated sentences) we conducted an initial MANOVA, with human ratings of affect categories the DVs (dependent variables) and the affect strength parameter $\beta$ the IV (independent variable). We then conducted follow-up univariate ANOVAs to identify which DV changes significantly with $\beta$. In total we conducted 5 MANOVAs and 30 follow-up ANOVAs, which required us to update the significance level to p$<$0.001 following a Bonferroni correction. 

\begin{figure*}[ht!]
 \centering
\captionsetup{justification=centering}
  \includegraphics[scale=0.53]{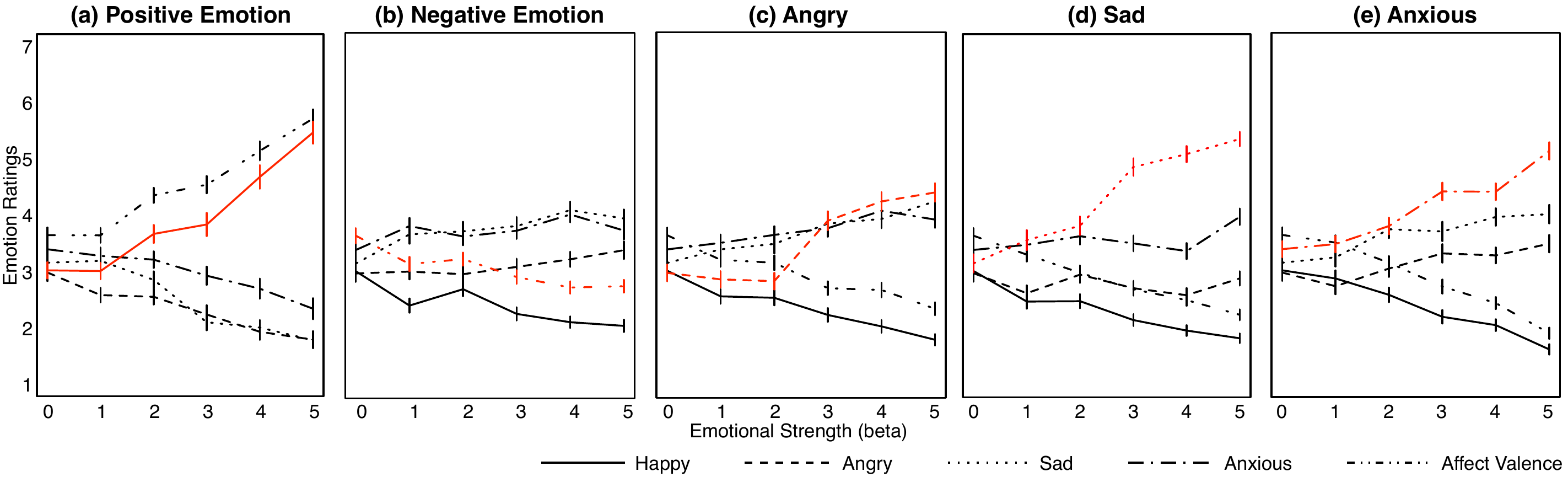}
\caption{Amazon Mechanical Turk study results for generated sentences in the target affect categories \emph{positive emotion, negative emotion, angry, sad}, and \emph{anxious} (a)-(e). The most relevant human rating curve for each generated emotion is highlighted in red, while less relevant rating curves are visualized in black. Affect categories are coded via different line types and listed in legend below figure.}
	\label{fig:mturk_experiment}
\end{figure*}

\section{Results}
\label{sec:results}
\subsection{Generation of Emotional Text}
In Section~\ref{subsec:generate-text} we have described the process of sampling text from the model conditioned on input affective information (research question \textbf{Q1}). Table~\ref{tab:sampling} shows three sentences generated by the model for input sentence beginnings \textit{I feel so ...}, \textit{Why did you ...} and \textit{I told him to ...} for each of five affect categories - \textit{happy(positive emotion), angry, sad anxiety}, and \textit{neutral(no emotion)}. They have been selected from a pool of 20 generated sentences for each category and sentence beginning. %\textit{Affect-LM} is trained on conversation transcripts and not grammatically vetted and correct news articles or Wikipedia entries, and thus there can be grammatical imperfections in the generated text. 
% Often these imperfections might follow the speaker's speaking manner as seen in the generated sentence \textit{I told him to be afraid of him and he he he just he just didn't care about the death penalty} from Table~\ref{tab:sampling} for the \textit{Anxious} category, which contains the repeated phrase \textit{he just}.
% Since the affect descriptors have been generated from the word context in a sequence-agnostic manner, there might be generated sentences which contain colored words without explicitly denoting the subject's emotional state. For example, \textit{I told him to see why he was fighting with my son} has been generated in the \textit{Anger} category due to the correlation of the word \textit{fighting} with \textit{Anger} as learnt by \textit{Affect-LM} from the training data. The generated sentence does not indicate that the subject (who speaks it) is angry himself.

\subsection{MTurk Perception Experiments}
\begin{figure}[thb]
 \centering
\captionsetup{justification=centering}
  \includegraphics[scale=0.50]{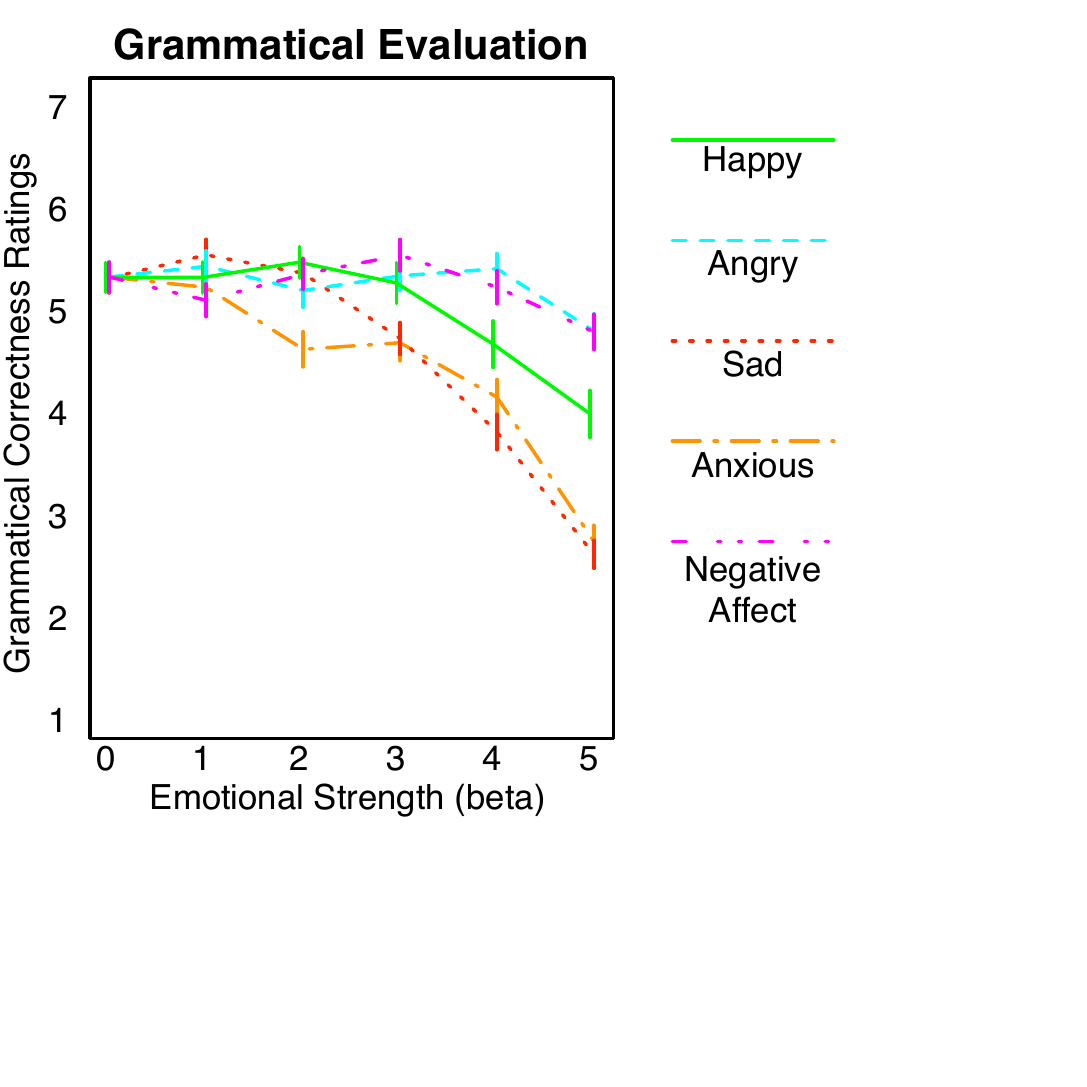}
  \caption{Mechanical Turk study results for \emph{grammatical correctness} for all generated target emotions. Perceived \emph{grammatical correctness} for each affect categories are color-coded.}
	\label{fig:grammatical_correctness}
\end{figure}

In the following we address research question \textbf{Q2} by reporting the main statistical findings of our MTurk study, which are visualized in Figures \ref{fig:mturk_experiment} and \ref{fig:grammatical_correctness}.

\textbf{Positive Emotion Sentences.} The multivariate result was significant for \emph{positive emotion} generated sentences (Pillai's Trace$=$.327, F(4,437)$=$6.44, p$<$.0001). Follow up ANOVAs revealed significant results for all DVs except \emph{angry} with p$<$.0001, indicating that both \emph{affective valence} and \emph{happy} DVs were successfully manipulated with $\beta$, as seen in Figure \ref{fig:mturk_experiment}(a). \emph{Grammatical correctness} was also significantly influenced by the affect strength parameter $\beta$ and results show that the correctness deteriorates with increasing $\beta$ (see Figure~\ref{fig:grammatical_correctness}). However, a post-hoc Tukey test revealed that only the highest $\beta$ value shows a significant drop in \emph{grammatical correctness} at p$<$.05.

\textbf{Negative Emotion Sentences.} The multivariate result was significant for \emph{negative emotion} generated sentences (Pillai's Trace$=$.130, F(4,413)$=$2.30, p$<$.0005). Follow up ANOVAs revealed significant results for \emph{affective valence} and \emph{happy} DVs with p$<$.0005, indicating that the \emph{affective valence} DV was successfully manipulated with $\beta$, as seen in Figure \ref{fig:mturk_experiment}(b). Further, as intended there were no significant differences for DVs \emph{angry, sad} and \emph{anxious}, indicating that the \emph{negative emotion} DV refers to a more general affect related concept rather than a specific negative emotion. This finding is in concordance with the intended LIWC category of \emph{negative affect} that forms a parent category above the more specific emotions, such as \emph{angry, sad}, and \emph{anxious} \cite{pennebaker2001linguistic}. \emph{Grammatical correctness} was also significantly influenced by the affect strength $\beta$ and results show that the correctness deteriorates with increasing $\beta$ (see Figure~\ref{fig:grammatical_correctness}). As for \emph{positive emotion}, a post-hoc Tukey test revealed that only the highest $\beta$ value shows a significant drop in \emph{grammatical correctness} at p$<$.05.

\textbf{Angry Sentences.} The multivariate result was significant for \emph{angry} generated sentences (Pillai's Trace$=$.199, F(4,433)$=$3.76, p$<$.0001). Follow up ANOVAs revealed significant results for \emph{affective valence, happy}, and \emph{angry} DVs with p$<$.0001, indicating that both \emph{affective valence} and \emph{angry} DVs were successfully manipulated with $\beta$, as seen in Figure \ref{fig:mturk_experiment}(c). \emph{Grammatical correctness} was not significantly influenced by the affect strength parameter $\beta$, which indicates that angry sentences are highly stable across a wide range of $\beta$ (see Figure~\ref{fig:grammatical_correctness}). However, it seems that human raters could not successfully distinguish between \emph{angry, sad}, and \emph{anxious} affect categories, indicating that the generated sentences likely follow a general \emph{negative affect} dimension.
\textbf{Sad Sentences.} The multivariate result was significant for \emph{sad} generated sentences (Pillai's Trace$=$.377, F(4,425)$=$7.33, p$<$.0001). Follow up ANOVAs revealed significant results only for the \emph{sad} DV with p$<$.0001, indicating that while the \emph{sad} DV can be successfully manipulated with $\beta$, as seen in Figure \ref{fig:mturk_experiment}(d). The \emph{grammatical correctness} deteriorates significantly with $\beta$. Specifically, a post-hoc Tukey test revealed that only the two highest $\beta$ values show a significant drop in \emph{grammatical correctness} at p$<$.05 (see Figure~\ref{fig:grammatical_correctness}). A post-hoc Tukey test for \emph{sad} reveals that $\beta=3$ is optimal for this DV, since it leads to a significant jump in the perceived sadness scores at p$<$.005 for $\beta \in \{0,1,2\}$.

\textbf{Anxious Sentences.} The multivariate result was significant for \emph{anxious} generated sentences (Pillai's Trace$=$.289, F(4,421)$=$6.44, p$<$.0001). Follow up ANOVAs revealed significant results for \emph{affective valence, happy} and \emph{anxious} DVs with p$<$.0001, indicating that both \emph{affective valence} and \emph{anxiety} DVs were successfully manipulated with $\beta$, as seen in Figure \ref{fig:mturk_experiment}(e). \emph{Grammatical correctness} was also significantly influenced by the affect strength parameter $\beta$ and results show that the correctness deteriorates with increasing $\beta$. Similarly for \emph{sad}, a post-hoc Tukey test revealed that only the two highest $\beta$ values show a significant drop in \emph{grammatical correctness} at p$<$.05 (see Figure~\ref{fig:grammatical_correctness}). Again, a post-hoc Tukey test for \emph{anxious} reveals that $\beta=3$ is optimal for this DV, since it leads to a 
\begin{table}[thb]
\centering
 \captionsetup{justification=centering}
\scriptsize
\begin{tabular}{|l|c|c|c|c|}
\hline
& \multicolumn{2}{|c|}{Training (Fisher)} & \multicolumn{2}{|c|}{Adaptation} \\ \hline
{\bf Perplexity} & {\bf Baseline}  & {\bf Affect-LM}  & {\bf Baseline} & {\bf Affect-LM}   \\\hline
Fisher & 37.97 &  37.89 & - &	- \\	\hline 
DAIC & 65.02 &  64.95 & 55.86 & 55.55 \\	\hline
SEMAINE & 88.18 &  86.12 & 57.58 & 57.26 \\ \hline
CMU-MOSI & 104.74 &  101.19 & 66.72 & 64.99 \\ \hline
Average & 73.98 &  72.54 & 60.05 & 59.26 \\ \hline
\end{tabular}
\caption{Evaluation perplexity scores obtained by the baseline and \textit{Affect-LM} models when trained on Fisher and subsequently adapted on DAIC, SEMAINE and CMU-MOSI corpora}
\label{fisher-perplex}
\end{table}
significant jump in the perceived anxiety scores at p$<$.005 for $\beta \in \{0,1,2\}$.

\subsection{Language Modeling Results}
In Table~\ref{fisher-perplex}, we address research question \textbf{Q3} by presenting the perplexity scores obtained by the baseline model and \textit{Affect-LM}, when trained on the Fisher corpus and subsequently adapted on three emotional corpora (each adapted model is individually trained on CMU-MOSI, DAIC and SEMAINE). The models trained on Fisher are evaluated on all corpora while each adapted model is evaluated only on it's respective corpus. For all corpora, we find that \textit{Affect-LM} achieves lower perplexity on average than the baseline model, implying that affect category information obtained from the context words improves language model prediction. The average perplexity improvement is 1.44 (relative improvement 1.94\%) for the model trained on Fisher, while it is 0.79 (1.31\%) for the adapted models. We note that larger improvements in perplexity are observed for corpora with higher content of emotional words. This is supported by the results in Table~\ref{fisher-perplex}, where \textit{Affect-LM} obtains a larger reduction in perplexity for the CMU-MOSI and SEMAINE corpora, which respectively consist of 2.76\% and 2.75\% more emotional words than the Fisher corpus.

\begin{figure}[bht]
 \centering
 \captionsetup{justification=centering}
  \includegraphics[scale=0.3]{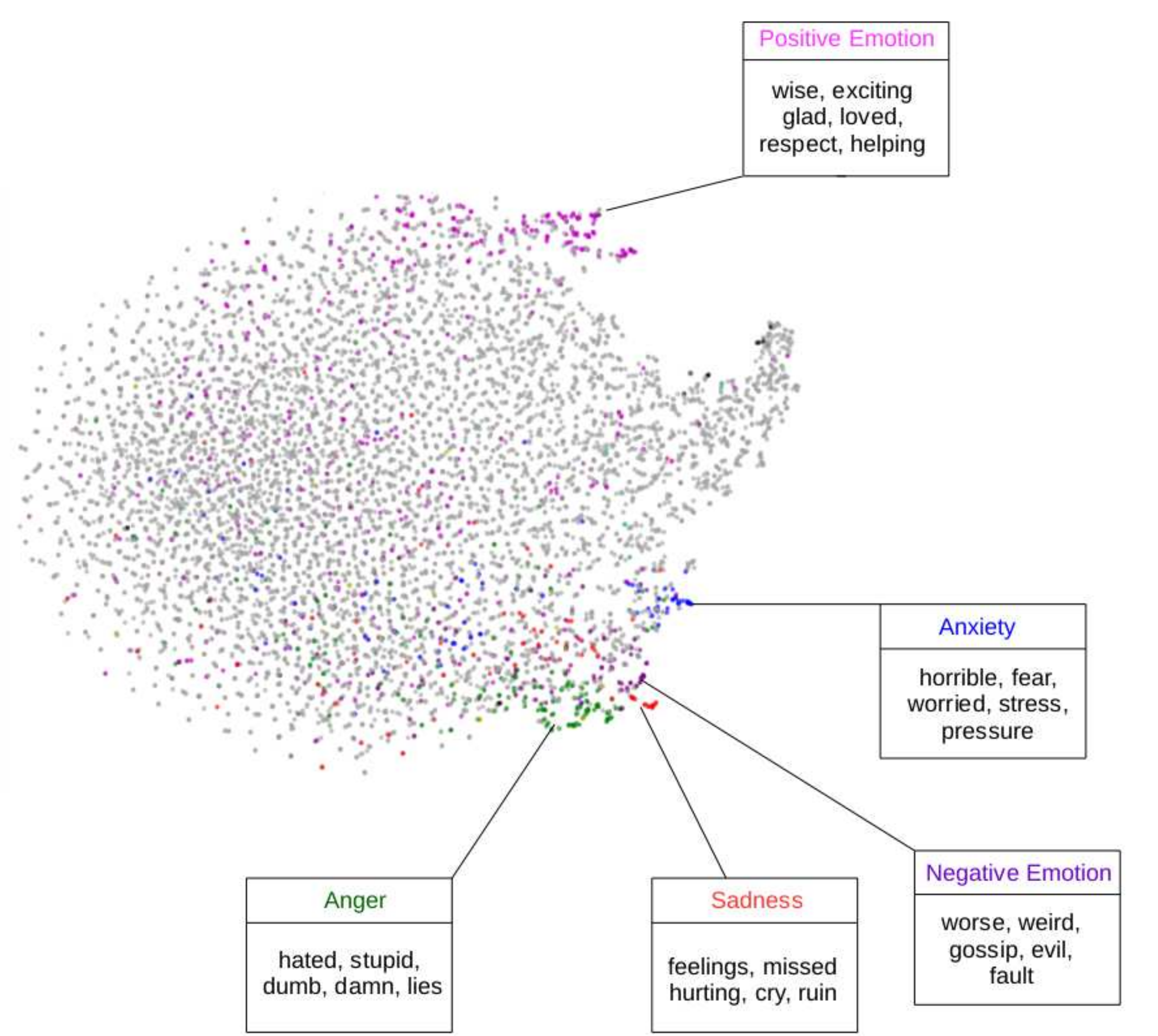}
  \caption{Embeddings learnt by \textit{Affect-LM}}
	\label{embed-repr}
\end{figure}

\subsection{Word Representations}
\label{subsec: word-repr}
In Equation~\ref{affectlm-eqn}, \textit{Affect-LM} learns a weight matrix $\mathbf{V}$ which captures the correlation between the predicted word $w_t$, and the affect category $\mathbf{e_{t-1}}$. Thus, each row of the matrix $\mathbf{V_i}$ is an emotionally meaningful embedding of the $i$-th word in the vocabulary. In Figure~\ref{embed-repr}, we present a visualization of these embeddings, where each data point is a separate word, and words which appear in the LIWC dictionary are colored based on which affect category they belong to (we have labeled only words in categories \textit{positive emotion}, \textit{negative emotion}, \textit{anger}, \textit{sad} and \textit{anxiety} since these categories contain the most frequent words). Words colored grey are those not in the LIWC dictionary. In Figure~\ref{embed-repr}, we observe that the embeddings contain affective information, where the positive emotion is highly separated from the negative emotions (\textit{sad}, \textit{angry}, \textit{anxiety}) which are clustered together.
%\vspace{-2.5mm}

\vspace{-2mm}
\section{Conclusions and Future Work}
\vspace{-1.5mm}
\label{sec:conclusions}
In this paper, we have introduced a novel language model~\textit{Affect-LM} for generating affective conversational text conditioned on context words, an affective category and an affective strength parameter. MTurk perception studies show that the model can generate expressive text at varying degrees of emotional strength without affecting grammatical correctness. %\textit{Affect-LM} utilizes the LIWC dictionary for feature extraction, and introduces an additional energy term for affective information in the model formulation.  
We also evaluate~\textit{Affect-LM} as a language model and show that it achieves lower perplexity than a baseline LSTM model when the affect category is obtained from the words in the context. For future work, we wish to extend this model by investigating language generation conditioned on other modalities such as facial images and speech, and to applications such as dialogue generation for virtual agents.
\vspace{-1mm}
\section*{Acknowledgments}
\vspace{-1mm}
This material is based upon work supported by the U.S. Army Research Laboratory under contract number W911NF-14-D-0005. Any opinions, findings, and conclusions or recommendations expressed in this material are those of the author(s) and do not necessarily reflect the views of the Government, and no official endorsement should be inferred. Sayan Ghosh also acknowledges the Viterbi Graduate School Fellowship for funding his graduate studies.

\bibliography{acl2017}

\begin{thebibliography}{}
\expandafter\ifx\csname natexlab\endcsname\relax\def\natexlab#1{#1}\fi

\bibitem[{Ahn et~al.(2016)Ahn, Choi, P{\"a}rnamaa, and Bengio}]{ahn2016neural}
Sungjin Ahn, Heeyoul Choi, Tanel P{\"a}rnamaa, and Yoshua Bengio. 2016.
\newblock A neural knowledge language model.
\newblock {\em arXiv preprint arXiv:1608.00318\/} .

\bibitem[{Bengio et~al.(2003)Bengio, Ducharme, Vincent, and
  Jauvin}]{bengio2003neural}
Yoshua Bengio, R{\'e}jean Ducharme, Pascal Vincent, and Christian Jauvin. 2003.
\newblock A neural probabilistic language model.
\newblock {\em Journal of machine learning research\/} 3(Feb):1137--1155.

\bibitem[{Buhrmester et~al.(2011)Buhrmester, Kwang, and
  Gosling}]{buhrmester2011amazon}
Michael Buhrmester, Tracy Kwang, and Samuel~D Gosling. 2011.
\newblock Amazon's mechanical turk a new source of inexpensive, yet
  high-quality, data?
\newblock {\em Perspectives on psychological science\/} 6(1):3--5.

\bibitem[{Bulyko et~al.(2007)Bulyko, Ostendorf, Siu, Ng, Stolcke, and
  {\c{C}}etin}]{bulyko2007web}
Ivan Bulyko, Mari Ostendorf, Manhung Siu, Tim Ng, Andreas Stolcke, and
  {\"O}zg{\"u}r {\c{C}}etin. 2007.
\newblock Web resources for language modeling in conversational speech
  recognition.
\newblock {\em ACM Transactions on Speech and Language Processing (TSLP)\/}
  5(1):1.

\bibitem[{Cho et~al.(2015)Cho, Courville, and Bengio}]{cho2015describing}
Kyunghyun Cho, Aaron Courville, and Yoshua Bengio. 2015.
\newblock Describing multimedia content using attention-based encoder-decoder
  networks.
\newblock {\em IEEE Transactions on Multimedia\/} 17(11):1875--1886.

\bibitem[{Cieri et~al.(2004)Cieri, Miller, and Walker}]{cieri2004fisher}
Christopher Cieri, David Miller, and Kevin Walker. 2004.
\newblock The fisher corpus: a resource for the next generations of
  speech-to-text.
\newblock In {\em LREC\/}. volume~4, pages 69--71.

\bibitem[{et~al.(2016)}]{abadi2016tensorflow}
Mart{\'\i}n~Abadi et~al. 2016.
\newblock Tensorflow: A system for large-scale machine learning.
\newblock In {\em Proceedings of the 12th USENIX Symposium on Operating Systems
  Design and Implementation (OSDI). Savannah, Georgia, USA\/}.

\bibitem[{Gratch(2014)}]{gratch2014distress}
Jonathan et~al. Gratch. 2014.
\newblock The distress analysis interview corpus of human and computer
  interviews.
\newblock In {\em LREC\/}. Citeseer, pages 3123--3128.

\bibitem[{Hochreiter and Schmidhuber(1997)}]{hochreiter1997long}
Sepp Hochreiter and J{\"u}rgen Schmidhuber. 1997.
\newblock Long short-term memory.
\newblock {\em Neural computation\/} 9(8):1735--1780.

\bibitem[{Jozefowicz et~al.(2016)Jozefowicz, Vinyals, Schuster, Shazeer, and
  Wu}]{jozefowicz2016exploring}
Rafal Jozefowicz, Oriol Vinyals, Mike Schuster, Noam Shazeer, and Yonghui Wu.
  2016.
\newblock Exploring the limits of language modeling.
\newblock {\em arXiv preprint arXiv:1602.02410\/} .

\bibitem[{Kao and Jurafsky(2012)}]{kao2012computational}
Justine Kao and Dan Jurafsky. 2012.
\newblock A computational analysis of style, affect, and imagery in
  contemporary poetry.

\bibitem[{Keshtkar and Inkpen(2011)}]{keshtkar2011pattern}
Fazel Keshtkar and Diana Inkpen. 2011.
\newblock A pattern-based model for generating text to express emotion.
\newblock In {\em Affective Computing and Intelligent Interaction\/}, Springer,
  pages 11--21.

\bibitem[{Kiros et~al.(2014)Kiros, Salakhutdinov, and
  Zemel}]{kiros2014multimodal}
Ryan Kiros, Ruslan Salakhutdinov, and Richard~S Zemel. 2014.
\newblock Multimodal neural language models.

\bibitem[{Mahamood and Reiter(2011)}]{mahamood2011generating}
Saad Mahamood and Ehud Reiter. 2011.
\newblock Generating affective natural language for parents of neonatal
  infants.
\newblock In {\em Proceedings of the 13th European Workshop on Natural Language
  Generation\/}. Association for Computational Linguistics, pages 12--21.

\bibitem[{Mairesse and Walker(2007)}]{mairesse2007personage}
Fran{\c{c}}ois Mairesse and Marilyn Walker. 2007.
\newblock Personage: Personality generation for dialogue.

\bibitem[{McKeown et~al.(2012)McKeown, Valstar, Cowie, Pantic, and
  Schroder}]{mckeown2012semaine}
Gary McKeown, Michel Valstar, Roddy Cowie, Maja Pantic, and Marc Schroder.
  2012.
\newblock The semaine database: Annotated multimodal records of emotionally
  colored conversations between a person and a limited agent.
\newblock {\em IEEE Transactions on Affective Computing\/} 3(1):5--17.

\bibitem[{Mikolov et~al.(2010)Mikolov, Karafi{\'a}t, Burget, Cernock{\`y}, and
  Khudanpur}]{mikolov2010recurrent}
Tomas Mikolov, Martin Karafi{\'a}t, Lukas Burget, Jan Cernock{\`y}, and Sanjeev
  Khudanpur. 2010.
\newblock Recurrent neural network based language model.
\newblock In {\em Interspeech\/}. volume~2, page~3.

\bibitem[{Mikolov et~al.(2013)Mikolov, Sutskever, Chen, Corrado, and
  Dean}]{mikolov2013distributed}
Tomas Mikolov, Ilya Sutskever, Kai Chen, Greg~S Corrado, and Jeff Dean. 2013.
\newblock Distributed representations of words and phrases and their
  compositionality.
\newblock In {\em Advances in neural information processing systems\/}. pages
  3111--3119.

\bibitem[{Nakov et~al.(2016)Nakov, Ritter, Rosenthal, Sebastiani, and
  Stoyanov}]{nakov2016semeval}
Preslav Nakov, Alan Ritter, Sara Rosenthal, Fabrizio Sebastiani, and Veselin
  Stoyanov. 2016.
\newblock Semeval-2016 task 4: Sentiment analysis in twitter.
\newblock {\em Proceedings of SemEval\/} pages 1--18.

\bibitem[{Pennebaker(2011)}]{pennebaker2011secret}
James~W Pennebaker. 2011.
\newblock The secret life of pronouns.
\newblock {\em New Scientist\/} 211(2828):42--45.

\bibitem[{Pennebaker et~al.(2001)Pennebaker, Francis, and
  Booth}]{pennebaker2001linguistic}
James~W Pennebaker, Martha~E Francis, and Roger~J Booth. 2001.
\newblock Linguistic inquiry and word count: Liwc 2001.
\newblock {\em Mahway: Lawrence Erlbaum Associates\/} 71(2001):2001.

\bibitem[{Picard(1997)}]{picard1997affective}
Rosalind Picard. 1997.
\newblock {\em Affective computing\/}, volume 252.
\newblock MIT press Cambridge.

\bibitem[{Scherer et~al.(2010)Scherer, B{\"a}nziger, and
  Roesch}]{scherer2010blueprint}
Klaus~R Scherer, Tanja B{\"a}nziger, and Etienne Roesch. 2010.
\newblock {\em A Blueprint for Affective Computing: A sourcebook and manual\/}.
\newblock Oxford University Press.

\bibitem[{Stolcke et~al.(2002)}]{stolcke2002srilm}
Andreas Stolcke et~al. 2002.
\newblock Srilm-an extensible language modeling toolkit.
\newblock In {\em Interspeech\/}. volume 2002, page 2002.

\bibitem[{Sundermeyer et~al.(2012)Sundermeyer, Schl{\"u}ter, and
  Ney}]{sundermeyer2012lstm}
Martin Sundermeyer, Ralf Schl{\"u}ter, and Hermann Ney. 2012.
\newblock Lstm neural networks for language modeling.
\newblock In {\em Interspeech\/}. pages 194--197.

\bibitem[{Vinyals et~al.(2015)Vinyals, Toshev, Bengio, and
  Erhan}]{Vinyals_2015_CVPR}
Oriol Vinyals, Alexander Toshev, Samy Bengio, and Dumitru Erhan. 2015.
\newblock Show and tell: A neural image caption generator.
\newblock In {\em The IEEE Conference on Computer Vision and Pattern
  Recognition (CVPR)\/}.

\bibitem[{Zadeh et~al.(2016)Zadeh, Zellers, Pincus, and
  Morency}]{zadeh2016multimodal}
Amir Zadeh, Rowan Zellers, Eli Pincus, and Louis-Philippe Morency. 2016.
\newblock Multimodal sentiment intensity analysis in videos: Facial gestures
  and verbal messages.
\newblock {\em IEEE Intelligent Systems\/} 31(6):82--88.

\bibitem[{Zaremba et~al.(2014)Zaremba, Sutskever, and
  Vinyals}]{zaremba2014recurrent}
Wojciech Zaremba, Ilya Sutskever, and Oriol Vinyals. 2014.
\newblock Recurrent neural network regularization.
\newblock {\em arXiv preprint arXiv:1409.2329\/} .

\end{thebibliography}
\bibliographystyle{acl_natbib}

\appendix
\end{document}